%% file: paper.tex
\documentclass{article}
\usepackage{kr}

\usepackage{times}
\usepackage{soul}
\usepackage{url}
\usepackage[hidelinks]{hyperref}
\usepackage[small]{caption}
\usepackage{graphicx}
\usepackage{amsmath}
\usepackage{amsthm}
\usepackage{booktabs}
\usepackage{algorithm}
\usepackage{algorithmicx}
\usepackage[noend]{algpseudocode}
\usepackage{graphicx}
\usepackage{microtype}
\usepackage{multirow}
\usepackage[dvipsnames]{xcolor}
\usepackage{etoolbox}
\usepackage{amssymb}
\usepackage{latexsym}
\usepackage{array}
\usepackage{adjustbox}
\usepackage{cite}
\usepackage{soul}
\usepackage{url}
\usepackage{fancybox}
\usepackage{subcaption}
\usepackage[english]{babel}
\usepackage{xspace}
\usepackage{setspace}
\usepackage{stmaryrd}
\usepackage{enumitem}
\usepackage{ifthen}
\usepackage{verbatim}
\usepackage{titlesec}
\usepackage{amsthm}
\usepackage[capitalise,nameinlink]{cleveref}

\usepackage[mode=text]{siunitx}
\usepackage{xfrac}
\usepackage{relsize}
\usepackage[semicolon]{natbib}

\usepackage{tikz}
\usepackage{forest}

\usetikzlibrary{arrows,decorations.pathmorphing,decorations.footprints,fadings,calc,trees,mindmap,shadows,decorations.text,patterns,positioning,shapes,matrix,fit}
\usetikzlibrary{arrows,shadows,backgrounds}
\usetikzlibrary{arrows.meta}
\usetikzlibrary{positioning,fit}
\usetikzlibrary{automata}
\usetikzlibrary{matrix}
\usetikzlibrary{shapes.symbols,shapes.misc,shapes.arrows,shapes.geometric}
\usetikzlibrary{matrix,chains,scopes,decorations.pathmorphing}

\usepackage{tikz-3dplot} 
\usepackage{pgfplots}
\pgfplotsset{compat=newest}

\newtheorem{example}{Example}

\input{defs}

\input{preamble}

\begin{document}

\maketitle

\input{abs}
\input{intro}
\input{prelim}

\input{lbxp}

\input{drxp}

\input{core}

\input{nualg}

\input{res}

\input{conc}
\input{ack}


\input{replbib}
\input{togbbl} 

\iftoggle{mkbbl}{
    \bibliographystyle{kr}
    \bibliography{refs,xrefs,xtra,team}
}{

\input{paper.bibl}
}

\clearpage
\appendix
\input{appdx}

\end{document}

%% file: defs.tex
\newboolean{nonanon}               
\setboolean{nonanon}{true}         %
\newboolean{mkcompact}             
\setboolean{mkcompact}{true}       %
\newboolean{squeezeit}             
\setboolean{squeezeit}{true}       %
\hypersetup{
  colorlinks = true,
  linkcolor = MidnightBlue,
  urlcolor  = MidnightBlue,
  citecolor = MidnightBlue,
  anchorcolor = MidnightBlue
}

\newtheorem{proposition}{Proposition}
\newtheorem{definition}{Definition}

\newtheorem{remark}{Remark}

\crefname{prop}{Proposition}{Propositions}
\Crefname{prop}{Proposition}{Propositions}
\crefname{defn}{Definition}{Definitions}
\Crefname{defn}{Definition}{Definitions}
\crefname{cor}{Corollary}{Corollaries}
\Crefname{cor}{Corollary}{Corollaries}
\crefname{exmpl}{Example}{Examples}
\Crefname{exmpl}{Example}{Examples}
\Crefname{theorem}{Theorem}{Theorem}

\newtheoremstyle{nutstyle}
{3.0pt} 
{3.0pt} 
{} 
{} 
{\bfseries} 
{.} 
{.5em} 
{} 

\theoremstyle{nutstyle}

\newtheoremstyle{nuxstyle}
{3.0pt} 
{3.0pt} 
{} 
{} 
{\itshape} 
{.} 
{.5em} 
{} 

\theoremstyle{nuxstyle}

\newcommand{\todoF}[2]{}

\newcommand{\fml}[1]{{\mathcal{#1}}}

\newcommand{\tn}[1]{\textnormal{#1}}
\newcommand{\mbf}[1]{\ensuremath\mathbf{#1}}
\newcommand{\msf}[1]{\ensuremath\mathsf{#1}}
\newcommand{\mbb}[1]{\ensuremath\mathbb{#1}}

\newcommand{\tbf}[1]{\textbf{#1}}

\newcommand{\pnorm}[1]{\ensuremath{l}_{#1}}
\newcommand{\refd}{\ensuremath\epsilon}
\newcommand{\refdg}{\ensuremath\eta}

\newcommand{\True}{\textbf{true}}
\newcommand{\False}{\textbf{false}}

\newcommand{\findaxpswift}{\ensuremath\msf{SwiftXplain}}
\newcommand{\findaxpdel}{\ensuremath\msf{FindAXpDel}}

\newcommand{\findaxpdicho}{\ensuremath\msf{FindAXpDichotomic}}
\newcommand{\qxprecur}{\ensuremath\msf{QxpRecur}}

\newcommand{\robt}{\ensuremath\msf{FindAdvEx}}

\newcommand{\outc}{\oper{hasAE}}

\newcommand{\swiftxp}{SwiftXplain\xspace}

\definecolor{darkslategray}{rgb}{0.18, 0.31, 0.31} 
\definecolor{platinum}{rgb}{0.9, 0.89, 0.89} 
\definecolor{gray}{rgb}{.4,.4,.4}
\definecolor{midgrey}{rgb}{0.5,0.5,0.5}
\definecolor{middarkgrey}{rgb}{0.35,0.35,0.35}
\definecolor{darkgrey}{rgb}{0.3,0.3,0.3}
\definecolor{darkred}{rgb}{0.7,0.1,0.1}
\definecolor{midblue}{rgb}{0.2,0.2,0.7}
\definecolor{darkblue}{rgb}{0.1,0.1,0.5}
\definecolor{defseagreen}{cmyk}{0.69,0,0.50,0}
\newcommand{\jnote}[1]{\medskip\noindent$\llbracket$\textcolor{darkred}{joao}: \emph{\textcolor{middarkgrey}{#1}}$\rrbracket$\medskip}

\newcommand{\jnoteF}[1]{}
\newcommand{\anoteF}[1]{}

\newcommand{\oper}[1]{\ensuremath\textnormal{\small{\textsf{#1}}}}

\newcounter{tableeqn}[table]

\DeclareMathOperator*{\limply}{\rightarrow}

\newcommand{\bigland}{\ensuremath\bigwedge}
\newcommand{\waxp}{\ensuremath\mathsf{WAXp}}
\newcommand{\wcxp}{\ensuremath\mathsf{WCXp}}
\newcommand{\axp}{\ensuremath\mathsf{AXp}}
\newcommand{\cxp}{\ensuremath\mathsf{CXp}}
\newcommand{\dsym}{\ensuremath\mathfrak{d}} 
\newcommand{\tdsym}{$\mathfrak{d}$} 
\newcommand{\ewaxp}{\ensuremath\dsym\;\!\!\mathsf{WAXp}}
\newcommand{\ewcxp}{\ensuremath\dsym\;\!\!\mathsf{WCXp}}
\newcommand{\eaxp}{\ensuremath\dsym\;\!\!\mathsf{AXp}}
\newcommand{\ecxp}{\ensuremath\dsym\mathsf{CXp}}
\newcommand{\aex}{\ensuremath\mathsf{AEx}}

\newcommand{\twaxp}{WAXp\xspace}

\newcommand{\tewaxp}{{\tdsym}\;\!\!WAXp\xspace}

\newcommand{\teaxp}{{\tdsym}\;\!\!AXp\xspace}
\newcommand{\tecxp}{{\tdsym}CXp\xspace}
\newcommand{\taex}{AEx\xspace}

\newcommand{\mailtodomain}[1]{\href{mailto:#1@icrea.cat}{\texttt{\nolinkurl{#1}}}}

\titleformat{\paragraph}[runin]
{\normalfont\bfseries}{}{0pt}{}

\newcolumntype{L}[1]{>{\raggedright\let\newline\\\arraybackslash\hspace{0pt}}m{#1}}
\newcolumntype{C}[1]{>{\centering\let\newline\\\arraybackslash\hspace{0pt}}m{#1}}
\newcolumntype{R}[1]{>{\raggedleft\let\newline\\\arraybackslash\hspace{0pt}}m{#1}}

\makeatletter
\newcommand\nparagraph{%
  \@startsection{paragraph}
    {4}
    {\z@}
    {0.225ex \@plus0.225ex \@minus.125ex}
    {-1em}
    {\normalfont\normalsize\bfseries}%
}
\makeatother

\titlespacing{\paragraph}{%
  0pt}{
  0.5\baselineskip}{
  1em}



%% file: preamble.tex
\title{Distance-Restricted Explanations:\\Theoretical Underpinnings \& Efficient Implementation}

\iftrue
\author{%
Yacine Izza$^\text{\normalfont 1}$\thanks{Corresponding Author. Email: izza@comp.nus.edu.sg}\and
Xuanxiang Huang$^\text{\normalfont 2}$\and
Antonio Morgado$^\text{\normalfont 3}$\\
Jordi Planes$^\text{\normalfont 4}$\and
Alexey Ignatiev$^\text{\normalfont 5}$\And
Joao Marques-Silva$^\text{\normalfont 6}$\\
\affiliations
$^\text{1}$CREATE, NUS, Singapore\and
$^\text{2}$CNRS@CREATE, Singapore\and
$^\text{3}$IST/INESC-ID, Portugal \\
$^\text{4}$University of Lleida, Spain \and
$^\text{5}$Monash University, Australia\and
$^\text{6}$ICREA, University of Lleida, Spain\\
}
\fi

%% file: abs.tex
\begin{abstract}
  The 
  uses of machine learning (ML) have snowballed in recent years.
  In many cases, ML models are highly complex, and their operation
  is beyond the understanding of human decision-makers.
  Nevertheless, some 
  uses of ML models involve high-stakes and safety-critical
  applications.
  Explainable artificial intelligence (XAI) aims to help human
  decision-makers in understanding the operation of such complex ML
  models, thus eliciting trust in their operation.
  Unfortunately, the majority of past XAI work is based on informal
  approaches, that offer no guarantees of rigor.
  Unsurprisingly,
  there exists comprehensive experimental and theoretical evidence
  confirming that informal methods of XAI can provide human-decision
  makers with erroneous information.
  Logic-based XAI represents a rigorous approach to
  explainability; it is model-based and offers the strongest
  guarantees of rigor of computed explanations.
  However, a well-known drawback of logic-based XAI is the complexity
  of logic reasoning, especially for highly complex ML models.
  Recent work proposed distance-restricted explanations,
  i.e.\ explanations that are rigorous provided the distance to a
  given input is small enough. 
  Distance-restricted explainability is tightly related with
  adversarial robustness, and it has been shown to scale for
  moderately complex ML models, but the number of inputs still
  represents a key limiting factor.
  This paper investigates novel algorithms for scaling up the
  performance of logic-based explainers when computing and enumerating
  ML model explanations 
  with a large number of inputs.
  %
\end{abstract}

%% file: intro.tex
\section{Introduction}
\label{sec:intro}

\jnoteF{%
  The advances in ML \& the need for trust. \\
  The importance of XAI and existing myths. \\
  The advances of logic-based explainability.\\
  The existing limitations and the insight of distance-restricted
  XPs.\\
  The paper's contributions.
}

Recent years have witnessed remarkable progress in machine learning
(ML).
In some domains, systems of ML far exceed human-level capabilities.
Motivated by an ever-increasing range of possible uses, the advances
in ML are having a profound societal impact, and that is expected to
continue in the foreseeable future. 
Nevertheless, trust in the operation of ML models is arguably the main
obstacle to their widespread deployment.
In application domains that directly affect human beings, the issue of
trust is paramount. These domains include those deemed of high-risk or
that are safety-critical.
Given the complexity of the most widely used ML models (including
highly complex neural networks (NNs)), one stepping stone for building
trust in the operation of ML models is to be able to explain the
operation of those models. This is the grand general objective of
eXplainable Artificial Intelligence (XAI).

Most work on XAI is based on informal methods, that offer no
guarantees of
rigor~(\citealp{guestrin-kdd16}; \citealp{lundberg-nips17};
\citealp{guestrin-aaai18}). The alternative is formal XAI, where
(local and/or global) explanations are represented by rigorous
logic-based definitions, which are then computed using practically
efficient automated reasoners~(\citealp{msi-aaai22};
\citealp{ms-rw22}).
However, formal XAI also exhibits a few crucial challenges, the most
visible of which being the complexity of reasoning, among
others~\citep{ms-rw22}.
Despite the challenges, there has been observable progress in formal
XAI since its inception in
2018/19~(\citealp{darwiche-ijcai18}; \citealp{inms-aaai19};
\citealp{ms-rw22}), including the ability for explaining complex tree 
ensembles~(\citealp{ims-ijcai21}; \citealp{iisms-aaai22};
\citealp{iisms-aaai24};). 
However, and until recently, NNs represented a major challenge for
formal XAI, with existing tools only capable of explaining very small
NNs, i.e.\ with a few tens of activation units~(\citealp{inms-aaai19};
\citealp{katz-fmcad23}; \citealp{katz-tacas23}).
Recent work~(\citealp{barrett-corr22}; \citealp{hms-corr23};
\citealp{barrett-nips23}) revisited a fundamental connection between
explanations and adversarial examples. (This connection had been proved
earlier~\citep{inms-nips19}, but in the more generalized context of
globally-defined explanations.)
By introducing the concept of distance-restricted
explanations~\citep{hms-corr23}, one is able to compute those
explanations using tools for finding adversarial examples. This result
is significant given the remarkable progress observed in such tools in
recent years~\citep{johnson-sttt23}.
Using highly optimized tools for deciding the existence of adversarial
examples, it is now possible to explain NNs with a few hundreds of
neurons~\citep{barrett-nips23}.

Despite this recent breakthrough, key challenges remain. The
algorithms used for computing abductive explanations mimic the
algorithms developed over the years for computing minimal
unsatisfiable subsets (MUSes) of logic formulas, which are also
referred to as minimal unsatisfiable cores (MUCs) in the case of
constraint programming~(\citealp{junker-aaai04};
\citealp{sais-ecai06}).
Well-known examples include the deletion-based
algorithm~\citep{chinneck-jc91}, the insertion-based
algorithm~\citep{puget-ecai88}, the Quickxplain
algorithm~\citep{junker-aaai04}, but also dichotomic
search~\citep{sais-ecai06} and the progression
algorithm~\citep{msjb-cav13}, among
others~(\citealp{msl-sat11}; \citealp{blms-aic12}).
More importantly, it is known that the same algorithms can also be
used for computing contrastive explanations, which mimic minimal
correction subsets (MCSes) of logic formulas, because in both cases
one is computing a minimal set over a monotone
predicate~\citep{msjm-aij17}.
Nevertheless, one fundamental limitation of such algorithms is that
they operate in a mostly sequential fashion, allowing little to no
flexibility for parallelization, i.e.\ constraints are excluded from a
minimal set one at a time. This fundamental limitation becomes more
acute in the case of abductive explanations when the number of
features is very large. For complex NNs, each call for deciding the
existence of an adversarial example can be time consuming. Running
thousands of calls in sequence becomes a major performance bottleneck.

This paper details novel insights towards parallelizing algorithms for
the computation of abductive explanations for ML models exhibiting a 
very large number of features.
Furthermore, the paper identifies novel key properties of
distance-restricted explanations, which allow not only the efficient
computation of distance-restricted abductive explanations, but also
the computation of distance-restricted contrastive explanations and
their enumeration.


%% file: prelim.tex
\section{Preliminaries}
\label{sec:prelim}

\jnoteF{%
  Distances.\\
  Classifiers.\\
  Adversarial examples.\\
}

\paragraph{Classification problems.}
%
Classification problems are defined on a set of features
$\fml{F}=\{1,\ldots,m\}$ and a set of classes
$\fml{K}=\{c_1,\ldots,c_K\}$.
Each feature $i$ has a domain $\mbb{D}_i$. Features can be ordinal or
categorical. Ordinal features can be discrete or real-valued.
Feature space is defined by the cartesian product of the features'
domains: $\mbb{F}=\mbb{D}_1\times\cdots\times\mbb{D}_m$.
A classifier computes a total function $\kappa:\mbb{F}\to\fml{K}$.
Throughout the paper, a classification problem $\fml{M}$ represents a
tuple $\fml{M}=(\fml{F},\mbb{F},\fml{K},\kappa)$.

An instance (or a sample) is a pair $(\mbf{v},c)$, with
$\mbf{v}\in\mbb{F}$ and $c\in\fml{K}$.
An explanation problem $\fml{E}$ is a tuple
$\fml{E}=(\fml{M},(\mbf{v},c))$. The generic purpose of XAI is to find
explanations for each given instance.
Moreover, when reasoning in terms of robustness, we are also
interested in the behavior of a classifier given some instances.
Hence, we will also use explanation problems when addressing
robustness.

\begin{example} \label{ex:runex}
  Throughout the paper, we consider the following classification
  problem.
  The features are $\fml{F}=\{1,2,3\}$, all ordinal with domains
  $\mbb{D}_1=\mbb{D}_2=\mbb{D}_3=\mbb{R}$. The set of classes is
  $\fml{K}=\{0,1\}$. Finally, the classification function is
  $\kappa:\mbb{F}\to\fml{K}$, defined as follows (with
  $\mbf{x}=(x_1,x_2,x_3)$): 
  \[
  \kappa(\mbf{x})=\left\{
  \begin{array}{lcl}
    1 & ~~ & \tn{if~} 0<{x_1}<2 \land 4x_1\ge(x_2+x_3) \\[3pt]
    0 & ~~ & \tn{otherwise}
  \end{array}
  \right.
  \]
  Moreover, let the target instance be $(\mbf{v},c)=((1,1,1),1)$.
\end{example}

\paragraph{Norm $\pnorm{p}$.}
%
The distance between two vectors $\mbf{v}$ and $\mbf{u}$ is denoted by
$\lVert\mbf{v}-\mbf{u}\rVert$, and the actual definition depends on
the norm being considered.
Different norms $\pnorm{p}$ can be considered.
For $p\ge1$, the $p$-norm is defined as follows~\citep{horn-bk12}:
\begin{equation}
  \begin{array}{lcl}
    \lVert\mbf{x}\rVert_{p} & {:=} &
    \left(\sum\nolimits_{i=1}^{m}|x_i|^{p}\right)^{\sfrac{1}{p}}
  \end{array}
\end{equation}

Let $d_i=1$ if $x_i\not=0$, and let $d_i=0$ otherwise. Then, for
$p=0$, we define the 0-norm, $\pnorm{0}$, as
follows~\citep{robinson-bk03}:
\begin{equation}
  \begin{array}{lcl}
    \lVert\mbf{x}\rVert_{0} & {:=} &
    \sum\nolimits_{i=1}^{m}d_i
  \end{array}
\end{equation}

In general, for $p\ge1$, $\pnorm{p}$ denotes the Minkowski distance.
Well-known special cases include
the Manhattan distance $\pnorm{1}$,
the Euclidean distance $\pnorm{2}$, and
the Chebyshev distance $\pnorm{\infty}=\lim_{p\to\infty}l_p$.
$\pnorm{0}$ denotes the Hamming distance.
In the remainder of the paper, we use $p\in\mbb{N}_0$ (but we also
allow $p=\infty$ for the Chebyshev distance).
%

\paragraph{Adversarial examples.}
%
Let $\fml{M}=(\fml{F},\mbb{F},\fml{K},\kappa)$ be a classification
problem. Let $(\mbf{v},c)$, with $\mbf{v}\in\mbb{F}$ and
$c=\kappa(\mbf{v})$, be a given instance. Finally, let $\refd>0$ be
a value of distance for norm $l_p$.

We say that there exists an adversarial example if the following logic
statement holds true,
\begin{equation} \label{eq:locrob}
  \exists(\mbf{x}\in\mbb{F}).\left(\lVert\mbf{x}-\mbf{v}\rVert_p\le\refd\right)\land\left(\kappa(\mbf{x})\not=c\right) 
\end{equation}
(The logic statement above holds true if there exists a point
$\mbf{x}$ which is less than $\refd$ distance (using norm $\pnorm{p}$)
from $\mbf{v}$, and such that the prediction changes.)
If~\eqref{eq:locrob} is false, then the classifier is said to be
$\refd$-robust.
If~\eqref{eq:locrob} is true, then any $\mbf{x}\in\mbb{F}$ for which
the following predicate holds:\footnote{Parameterizations are shown as
predicate arguments positioned after ';'. These may be dropped for the
sake of brevity.}
%
\begin{equation} \label{eq:ae}
  \aex(\mbf{x};\fml{E},\refd,p) ~:=~
  \left(\lVert\mbf{x}-\mbf{v}\rVert_p\le\refd\right)\land\left(\kappa(\mbf{x})\not=c\right) 
\end{equation}
is referred to as an \emph{adversarial example}.%
\footnote{%
For regression problems~\citep{barrett-nips23} considers two
distance-related parameters for characterizing adversarial examples:
i) the distance to the point being considered, which corresponds to
$\epsilon$ in the definitions above; and ii) the amount of change in
the output that is deemed relevant, which corresponds to a maximum
value $\delta>0$ between the prediction of $\mbf{v}$ and any other
point restricted by the distance $\epsilon$. Throughout this paper, we
will only study classification problems; the case of regression
problems could adopt the definitions from earlier
work~\citep{barrett-nips23}.}
Tools that decide the existence of adversarial examples will be
referred to as \emph{robustness tools}. (In the case of neural
networks (NNs), the progress observed in robustness tools is
documented by VNN COMP~\citep{johnson-sttt23}.)

\begin{example} \label{ex:runex:ae}
  For the classifier from~\cref{ex:runex}, for distance $l_1$, and
  with $\refd=1$, there exist adversarial examples by either setting
  $x_1=0$ or $x_1=2$.
\end{example}

\jnoteF{Introduce here the concept of constrained AXps.}

It may happen that we are only interested in inputs that respect some
constraint, i.e.\ not all points of feature space are allowed or 
interesting. In such cases, we define adversarial examples subject to
some constraint $\fml{C}:\mbb{F}\to\{0,1\}$, which are referred to as
\emph{constrained AExs}. In this case, the adversarial examples must
satisfy the following logic statement:
\begin{equation} \label{eq:ae2}
  \fml{C}(\mbf{x})\land\left(\lVert\mbf{x}-\mbf{v}\rVert_p\le\refd\right)\land\left(\kappa(\mbf{x})\not=c\right)
\end{equation}
Clearly, the predicate $\aex$ (see~\eqref{eq:ae}) can be parameterized
by the constraint being used.

\paragraph{Minimal hitting sets.}
%
Let $\fml{S}$ be a set and $\mbb{B}\subseteq2^{\fml{S}}$ be a set of
subsets of $\fml{S}$. A hitting set (HS) $\fml{H}\subseteq\fml{S}$ of
$\mbb{B}$ is such that
$\forall(\fml{P}\in\mbb{B}).\fml{P}\cap\fml{H}\not=\emptyset$. A
minimal hitting set (MHS) $\fml{Q}\subseteq\fml{S}$ is a hitting set
of $\mbb{B}$ such that no proper subset of $\fml{Q}$ is a hitting
set of $\mbb{B}$,
i.e.\ $\forall(\fml{R}\subsetneq\fml{Q})\exists(\fml{P}\in\mbb{B}).\fml{R}\cap\fml{P}=\emptyset$.
A minimal hitting set is said to be subset-minimal or irreducible.

\paragraph{MUSes, MCSes, etc.}
The paper assumes basic knowledge of minimal unsatisfiable subsets
(MUSes), minimal correction subsets (MCSes) and related
concepts in the context of logic formulas. Examples of recent
accounts of these concepts
include~(\citealp{msjm-aij17}; \citealp{msm-ijcai20};
\citealp{sat-handbook21}; \citealp{osullivan-ijcai21}).

%% file: lbxp.tex
\section{Logic-Based Explainability}
\label{sec:lbxp}

\jnoteF{%
  Brief overview of logic-based explainability.\\
  AXps vs.\ CXps.\\
  Duality.\\
  Algorithms: akin to MUS extraction.\\
  The existing limitations of logic-based XAI.
}

%
In the context of explaining ML models, rigorous, model-based
explanations have been studied since 2018~\citep{darwiche-ijcai18}. We
follow recent treatments of the
subject~(\citealp{msi-aaai22}; \citealp{ms-rw22}).%
\footnote{%
Alternative accounts, using somewhat different notations, are also
available~(\citealp{marquis-kr21}; \citealp{darwiche-jlli23};
\citealp{darwiche-lics23}).}
%
A PI-explanation (which is also referred to as an abductive
explanation (AXp)~\citep{msi-aaai22}) is an irreducible subset of the
features which, if fixed to the values specified by a given instance,
are sufficient for the prediction.

Given an instance $(\mbf{v},c)$, a set of features
$\fml{X}\subseteq\fml{F}$ is sufficient for the prediction if the
following logic statement holds true,
\begin{equation} \label{eq:waxp}
  \forall(\mbf{x}\in\mbb{F}).\left[\land_{i\in\fml{X}}(x_i=v_i)\right]\limply\left(\kappa(\mbf{x})=c)\right) 
\end{equation}
If~\eqref{eq:waxp} holds, but $\fml{X}$ is not necessarily irreducible
(i.e.\ it is not subset-minimal), then we say that $\fml{X}$ is a weak
abductive explanation (WAXp). As a result, we associate a predicate
$\waxp$ with~\eqref{eq:waxp}, such that $\waxp(\fml{X};\fml{E})$ holds
true iff~\eqref{eq:waxp} holds true. An AXp is a weak AXp that is
subset-minimal. The predicate $\axp(\fml{X};\refd,\fml{E})$ holds true iff
set $\fml{X}$ is also an AXp. An AXp answers a \emph{Why?} question,
i.e.\ why is the prediction $c$ (given the values assigned to the
features).

A set of features $\fml{Y}\subseteq\fml{F}$ is sufficient for changing
the prediction if the following logic statement holds true,
\begin{equation} \label{eq:wcxp}
  \exists(\mbf{x}\in\mbb{F}).\left[\land_{i\in\fml{F}\setminus\fml{Y}}(x_i=v_i)\right]\land\left(\kappa(\mbf{x})\not=c)\right) 
\end{equation}
If~\eqref{eq:wcxp} holds, but $\fml{Y}$ is not necessarily
irreducible, then we say that $\fml{Y}$ is a weak contrastive
explanation (CXp). As a result, we associate a predicate $\wcxp$
with~\eqref{eq:wcxp}, such that $\wcxp(\fml{Y};\fml{E})$ holds true 
iff~\eqref{eq:wcxp} holds true. A CXp is a weak CXp that is also
subset-minimal. The predicate $\cxp(\fml{Y};\fml{E})$ holds true iff
set $\fml{Y}$ is a CXp. A CXp can be viewed as answering a
\emph{Why~Not?} question~(\citealp{miller-aij19};
\citealp{inams-aiia20}), i.e.\ why not some prediction other than $c$
(given the values assigned to the features).
Given an explanation problem, if at least one of the features in each
CXp is not allowed to change value, then the prediction remains
unchanged.

\begin{example}
  For the classifier of~\cref{ex:runex}, and given the target instance,
  the AXp is $\{1,2,3\}$. Clearly, if we allow any feature to take any
  value, then we can change the prediction. Hence, the prediction does
  not change only if all features are fixed.
\end{example}

Given the above, we define,
\begin{align}
  \mbb{A}(\fml{E}) &= \{\fml{X}\subseteq\fml{F}\,|\,\axp(\fml{X};\fml{E})\}
  \label{eq:allaxp}\\
  \mbb{C}(\fml{E}) &= \{\fml{Y}\subseteq\fml{F}\,|\,\cxp(\fml{Y};\fml{E})\}
  \label{eq:allcxp} 
\end{align}
which capture, respectively, the set of all AXps and the set of all
CXps given an explanation problem $\fml{E}$.
We define related sets for weak AXps and CXps, as follows:
\begin{align}
  \mbb{WA}(\fml{E}) &= \{\fml{X}\subseteq\fml{F}\,|\,\waxp(\fml{X};\fml{E})\}
  \label{eq:allwaxp}\\
  \mbb{WC}(\fml{E}) &= \{\fml{Y}\subseteq\fml{F}\,|\,\wcxp(\fml{Y};\fml{E})\}
  \label{eq:allwcxp} 
\end{align}

A simple observation is that WAXps are hitting sets of the sets of
WCXps, and WCXps are hitting sets of the sets of WAXps. For example, if
some WAXp $\fml{X}$ were not a hitting set of the set of WCXps, then
there would exist some non-hit set $\fml{Y}$ that would enable
changing the prediction whereas the disjoint set $\fml{X}$ would
enable fixing the prediction; a contradiction.
The previous observation can be substantially refined. The following
result relating AXps and CXps is used extensively in devising
explainability algorithms~\citep{inams-aiia20}.
\begin{proposition}[MHS Duality between AXps and CXps] \label{prop:duality1}
  Given an explanation problem $\fml{E}$,
  \begin{enumerate}[nosep]
  \item A set $\fml{X}\subseteq\fml{F}$ is an AXp iff $\fml{X}$ a
    minimal hitting set of the CXps in $\mbb{C}(\fml{E})$.
  \item A set $\fml{Y}\subseteq\fml{F}$ is a CXp iff $\fml{Y}$ a
    minimal hitting set of the AXps in $\mbb{A}(\fml{E})$.
  \end{enumerate}
\end{proposition}
(\cref{prop:duality1} is a consequence of an earlier seminal result in
model-based diagnosis~\citep{reiter-aij87}.)
\cref{prop:duality1} is instrumental for enumerating abductive (but
also contrastive) explanations~\citep{inams-aiia20}.
In contrast with non-formal explainability, the navigation of the
space of abductive (or contrastive) explanations, i.e.\ their
enumeration, is a hallmark of formal
XAI~(\citealp{inams-aiia20}; \citealp{marquis-kr21};
\citealp{msi-aaai22}). 

\jnoteF{To
  cite:~\citep{darwiche-jlli23,marquis-dke22,msi-aaai22,ms-corr22}.}

%% file: drxp.tex
\section{Distance-Restricted Explainability}
\label{sec:drxp}

\jnoteF{%
  Introduce distance-restricted XPs.\\
  dAXps vs.\ dCXps.\\
  Duality.\\
  Additional results: distance-minimal adversarial examples correspond
  CXps.\\
  Identify limitations of existing algorithms, namely VeriX.\\
  Relate with MUS extraction, and its known limitations.
}

\jnoteF{Put the proofs in appendix!!!}

\jnoteF{Relate with Adversarial Examples!!!}

This section proposes a generalized definition of (W)AXps and (W)CXps,
that take into account the $\pnorm{p}$ distance between $\mbf{v}$ and
the points that can be considered in terms of changing the prediction
$c=\kappa(\mbf{v})$.%
\footnote{%
\cref{sec:drxp,sec:core} substantially extend preliminary ideas
contained in an earlier report~\citep{hms-corr23}.}
%
%
The section starts by defining distance-restricted AXps/CXps,
i.e.\ {\teaxp}a/{\tecxp}s, which take the $\pnorm{p}$ distance into
account. Afterwards, the section proves a number of properties
motivated by the proposed generalized definition of AXps \& CXps,
including that MHS duality between AXps and CXps extends to the
distance-restricted definition of explanations.

\paragraph{Definitions.}
%
The standard definitions of AXps \& CXps can be generalized to take a
measure $\pnorm{p}$ of distance into account.

\begin{definition}[Distance-restricted (W)AXp, \tdsym(W)AXp]
  For a norm $\pnorm{p}$, a set of features $\fml{X}\subseteq\fml{F}$
  is a distance-restricted weak abductive explanation (\twaxp) for an
  instance $(\mbf{v},c)$, within distance $\refd>0$ of $\mbf{v}$, if
  the following predicate holds true,
  \begin{align} \label{eq:waxpg}
    &\ewaxp(\fml{X};\fml{E},\refd,p) ~:=~ \forall(\mbf{x}\in\mbb{F}).\\
    & \left(\bigwedge\nolimits_{i\in\fml{X}}(x_i=v_i)\land(\lVert\mbf{x}-\mbf{v}\rVert_{p}\le\refd)\right)\limply(\kappa(\mbf{x})=c)\nonumber
  \end{align}
  If a (distance-restricted) weak AXp $\fml{X}$ is irreducible
  (i.e.\ it is subset-minimal), then $\fml{X}$ is a
  (distance-restricted) AXp, or \teaxp.
\end{definition}

\begin{definition}[Distance-restricted (W)CXp, \tdsym(W)CXp]
  For a norm $\pnorm{p}$, a set of features $\fml{Y}\subseteq\fml{F}$
  is a weak contrastive explanation (WCXp) for an instance
  $(\mbf{v},c)$, within distance $\refd>0$ of $\mbf{v}$, if the
  following predicate holds true,
  \begin{align} \label{eq:wcxpg}
    &\ewcxp(\fml{Y};\fml{E},\refd,p) ~:=~ \exists(\mbf{x}\in\mbb{F}). \\
    &\left(\bigwedge\nolimits_{i\in\fml{F}\setminus\fml{Y}}(x_i=v_i)\land(\lVert\mbf{x}-\mbf{v}\rVert_{p}\le\refd)\right)\land(\kappa(\mbf{x})\not=c) \nonumber
  \end{align}
  If a (distance-restricted) weak CXp $\fml{Y}$ is irreducible,
  then $\fml{Y}$ is a (distance-restricted) CXp, or \tecxp.
\end{definition}

Furthermore, when referring to {\teaxp}s (resp.~{\tecxp}s), the
predicates $\eaxp$ (resp.~$\ecxp$) will be used.

\begin{example}
  For the classifier of~\cref{ex:runex}, let the norm used be $l_1$,
  with distance value $\refd=1$. From~\cref{ex:runex:ae}, we know that
  there exist adversarial examples, e.g.\ by setting $x_1=0$ or
  $x_1=2$. However, if we fix the value of $x_1$ to 1, then any
  assignment to $x_2$ and $x_3$ with $|x_2-1|+|x_3-1|\le1$, will not
  change the prediction. As a result, $\fml{X}=\{1\}$ is a
  distance-restricted AXp when $\refd=1$. Moreover, by allowing only
  feature 1 to change value, we are able to change prediction, since
  we know there exists an adversarial example.
\end{example}


The (distance-unrestricted) AXps (resp.~CXps) studied in earlier
work~(\citealp{darwiche-ijcai18}; \citealp{inms-aaai19}) represent a 
specific case of the distance-restricted AXps (resp.~CXps) introduced
in this section.

\begin{remark}
  Distance unrestricted AXps (resp.~CXps) correspond to
  $m$-distance {\teaxp}s (resp.~{\tecxp}s) for norm $l_0$, where $m$
  is the number of features.
\end{remark}

\paragraph{Properties.}
%
%
Distance-restricted explanations exhibit a number of relevant
properties.

The following observation will prove useful in designing efficient
algorithms (on the complexity of the oracle for adversarial examples)
for finding distance-restricted AXps/CXps.
\begin{proposition} \label{prop:monoent}
  The predicates $\ewaxp$ and $\ewcxp$ are monotonically increasing
  with respect to set inclusion (i.e.\ they are both monotonic and
  up-closed).
\end{proposition}
\cref{prop:monoent} mimics a similar earlier observation for
$\waxp$ and $\wcxp$
(e.g.\ see~\citep{inams-aiia20}), and follows from monotonicity of
entailment.

Moreover, it is apparent that $\refd$AXps and $\refd$CXps offer
a rigorous definition of the concept of \emph{local} explanations
studied in non-formal XAI~\citep{molnar-bk20}.

\jnoteF{There exists a (non-empty) CXp iff there exists an AEx. There
  exists a (non-empty) AXp iff there exists an AEx.}

\begin{proposition} \label{prop:aex2xp}
  Consider an explanation problem $\fml{E}=(\fml{M},(\mbf{v},c))$ and
  some $\refd>0$ for norm $\pnorm{p}$. Let $\mbf{x}\in\mbb{F}$, with
  $\lVert\mbf{x}-\mbf{v}\rVert_p\le\refd$, and let
  $\fml{D}=\{i\in\fml{F}\,|\,x_i\not=v_i\}$. 
  Then,
  \begin{enumerate}[nosep]
  \item If $\aex(\mbf{x};\fml{E},\refd,p)$ holds, then
    $\ewcxp(\fml{D};\fml{E},\refd,p)$
    holds;
  \item If $\ewcxp(\fml{D};\fml{E},\refd,p)$
    holds, then
    $\exists(\mbf{y}\in\mbb{F}).
    \lVert\mbf{y}-\mbf{v}||_p\le||\mbf{x}-\mbf{v}\rVert_p\land
    \aex(\mbf{y};\fml{E},\refd,p)$.
  \end{enumerate}
\end{proposition}
  
\begin{proof}[Proof sketch]
  We prove each claim separately.
  \begin{enumerate}[nosep]
  \item This case is immediate. Given $\refd$ and $\fml{E}$, the
    existence of an adversarial example guarantees, by~\eqref{eq:wcxpg},
    the existence of a distance-restricted
    weak CXp.
  \item By~\eqref{eq:wcxpg}, if the features in
    $\fml{D}$ are allowed to change, given the distance restriction,
    then there exists at least one point $\mbf{y}$ such that the
    prediction changes for $\mbf{y}$, and the distance from $\mbf{y}$
    to $\mbf{v}$ does not exceed that from $\mbf{x}$ to $\mbf{v}$.
    \qedhere
  \end{enumerate}
\end{proof}

Given the definitions above, we generalize~\eqref{eq:allaxp}
and~\eqref{eq:allcxp} to distance-restricted explanations, as follows:
\begin{align}
  \dsym\mbb{A}(\fml{E},\refd;p) &= \{\fml{X}\subseteq\fml{F}\,|\,\eaxp(\fml{X};\fml{E},\refd,p)\}
  \label{eq:allaxpg}\\
  \dsym\mbb{C}(\fml{E},\refd;p) &= \{\fml{Y}\subseteq\fml{F}\,|\,\ecxp(\fml{Y};\fml{E},\refd,p)\}
  \label{eq:allcxpg} 
\end{align}

As before, we define related sets for weak {\teaxp}s and {\tecxp}s, as
follows:
\begin{align}
  \dsym\mbb{WA}(\fml{E},\refd;p) &= \{\fml{X}\subseteq\fml{F}\,|\,\ewaxp(\fml{X};\fml{E},\epsilon,p)\}
  \label{eq:allwaxpg}\\
  \dsym\mbb{WC}(\fml{E},\refd;p) &= \{\fml{Y}\subseteq\fml{F}\,|\,\ewcxp(\fml{Y};\fml{E},\epsilon,p)\}
  \label{eq:allwcxpg} 
\end{align}


In turn, this yields the following result regarding minimal hitting
set duality between \tdsym(W)AXps and \tdsym(W)CXps.

\begin{proposition} \label{prop:duality2}
  Given an explanation problem $\fml{E}$, norm $\pnorm{p}$, and a
  value of distance $\refd>0$ then,
  \begin{enumerate}[nosep]
  \item A set $\fml{X}\subseteq\fml{F}$ is a \teaxp iff $\fml{X}$ is a
    minimal hitting set of the {\tecxp}s in
    $\dsym\mbb{C}(\fml{E},\refd;p)$.
  \item A set $\fml{Y}\subseteq\fml{F}$ is a \tecxp iff $\fml{Y}$ is a
    minimal hitting set of the {\teaxp}s in
    $\dsym\mbb{A}(\fml{E},\refd;p)$.
  \end{enumerate}
\end{proposition}
    

\cref{prop:duality2} is instrumental for the enumeration of {\teaxp}s
and {\tecxp}s, as shown in earlier work in the case of
distance-unrestricted AXps/CXps~\citep{inams-aiia20}, since it enables
adapting well-known algorithms for the enumeration of subset-minimal
reasons of inconsistency~\citep{lpmms-cj16}.

\begin{example}
  For the running example, we have that
  $\dsym\mbb{A}(\fml{E},1;1)=\dsym\mbb{C}(\fml{E},1;1)=\{\{1\}\}$.
\end{example}

The definitions of distance-restricted AXps and CXps also reveal
novel uses for abductive \& contrastive explanations.
For a given distance $\refd$, a \teaxp represents an irreducible
sufficient reason for the ML model not to have an adversarial
example.

The number of distance-restricted WAXps/WCXps is non-decreasing with
the distance $\refd$.

\begin{proposition}
\label{prop:dwaxp_epsilon_eta}
  Let $\fml{E}$ represent some explanation problem.
  Let $0<\epsilon<\eta$ and $p$ some norm. For any
  $\fml{S}\subseteq\fml{F}$:
  \begin{enumerate}[nosep]
  \item If $\ewaxp(\fml{S};\fml{E},\eta,p)$, then
    $\ewaxp(\fml{S};\fml{E},\epsilon,p)$;
  \item If $\ewcxp(\fml{S};\fml{E},\epsilon,p)$, then
    $\ewcxp(\fml{S};\fml{E},\eta,p)$;
  \end{enumerate}
\end{proposition}

A similar result does not hold for {\teaxp}s/{\tecxp}s, as the
following examples illustrate:
\begin{example}
Consider a classifier $\fml{M}$ defined on $\fml{F} = \{1, 2\}$ and $\fml{K}=\{0, 1\}$,
with feature domains $\mbb{D}_1=\{0,0.5,1\}$ and $\mbb{D}_2=\{0,0.5,1\}$.
Let $\kappa(x_1,x_2)$ be its classification function
such that $\kappa(0.5,0.5) = 0$, $\kappa(0,1) = 0$, $\kappa(1,0) = 0$.
For any other point $\mbf{x}$, we have $\kappa(\mbf{x})=1$.
Let $\fml{E} = (\fml{M}, ((1,1),1))$.
Let us use $l_\infty$, and suppose $\epsilon_1 = 0.5$ and $\epsilon_2 = 1$.
For $\epsilon_1$, there is one AEx $\{(0.5,0.5)\}$, from which we deduce
$\dsym\mbb{C}(\fml{E},\refd_1;\infty) = \{\{1,2\}\}$.
For $\epsilon_2$, there are three AEx $\{(0.5,0.5), (1,0), (0,1)\}$, but then we can deduce
$\dsym\mbb{C}(\fml{E},\refd_2;\infty) = \{\{1\}, \{2\}\}$.
In this case, $\{1,2\} \not\in \dsym\mbb{C}(\fml{E},\refd_2;\infty)$.
\end{example}

\begin{example}
Consider a classifier $\fml{M}$ defined on $\fml{F} = \{1, 2, 3\}$ and $\fml{K}=\{0, 1\}$,
with feature domains $\mbb{D}_1=\mbb{D}_3=\{-0.5,0,0.5,1\}$, and $\mbb{D}_2=\{0,0.5,1\}$.
Let $\kappa(x_1,x_2,x_3)$ be its classification function
such that $\kappa(0.5,0.5,1) = 0$, $\kappa(1,0.5,0.5) = 0$, $\kappa(-0.5,1,1) = 0$ and $\kappa(1,1,-0.5) = 0$.
For any other point $\mbf{x}$, we have $\kappa(\mbf{x})=1$.
Let $\fml{E} = (\fml{M}, ((1,1,1),1))$.
Let us use $l_1$, and suppose $\epsilon_1 = 1$ and $\epsilon_2 = 1.5$.
For $\epsilon_1=1$, there are two AEx $\{(0.5,0.5,1), (1,0.5,0.5)\}$, from which we deduce
$\dsym\mbb{C}(\fml{E},\refd_1;1) = \{\{1,2\}, \{2,3\}\}$.
For $\epsilon_2=1.5$, there are four AEx $\{(0.5,0.5,1), (1,0.5,0.5), (-0.5,1,1), (1,1,-0.5)\}$, but then we can deduce
$\dsym\mbb{C}(\fml{E},\refd_2;1) = \{\{1\}, \{3\}\}$.
By MHS, we have $\dsym\mbb{A}(\fml{E},\refd_1;1) = \{\{2\}, \{1,3\}\}$,
but $\dsym\mbb{A}(\fml{E},\refd_2;1) = \{\{1,3\}\}$.
Clearly, $\{2\} \not \in \dsym\mbb{A}(\fml{E},\refd_2;1)$.
\end{example}

%

\paragraph{Related work on distance-restricted explanations.}
Distance-restricted explanations have been studied in recent
works~(\citealp{barrett-corr22}; \citealp{hms-corr23};
\citealp{barrett-nips23}).
The tight relationship between distance-restricted AXps and
distance-unrestricted AXps is first discussed in~\citep{hms-corr23};
other works~(\citealp{barrett-corr22}; \citealp{barrett-nips23}) do
not address this relationship.
%
%
%
Furthermore, some properties of distance-restricted explanations can
be related with those studied in earlier work on computing
explanations subject to additional constraints~(\citealp{cms-cp21}; 
\citealp{cms-aij23})
or
subject to background knowledge~(\citealp{rubin-aaai22};
\citealp{yisnms-aaai23}), including hitting-set duality.

\jnoteF{%
  Link betwen recent work on distance-restricted explanations and AXps
  (or PI-explanations) is not explicit, e.g.~\citep{barrett-nips23};
  This paper formalizes this relationship; in turn this is crucial for
  devising algorithms for navigating the space of distance-restricted
  explanations.\\
  Constrained AExs, adapted from constrained AXps/CXps~\citep{cms-aij23};\\
  How to relate $\ewaxp$s and $\ewcxp$s with constrained AExs;\\
}

\jnoteF{%
  Additional remarks:
  \begin{enumerate}[nosep]
  \item For $l_0$, only {\tecxp}s respect $|\fml{Y}|\le\epsilon$.
  \item {\teaxp}s are only guaranteed to be minimal hitting sets of
    the {\tecxp}s. 
  \end{enumerate}
}

%% file: core.tex
\section{Explanations Using Adversarial Examples} \label{sec:core}

\jnoteF{%
  Overview sequential algorithms for finding one AXp, one CXp and for
  their enumeration. \\
  Recap MSMP. \\
  Cover deletion, dichotomic, progression, quickxplain. \\
  Analyze limitations of algorithms for finding on AXp.
}

\jnoteF{%
  Select 1 or 2 algorithms to describe. Any other algorithms will be
  moved to appendix.
}

This section shows that the formal framework that has been developed
in the case of computing and enumerating AXps \& CXps can be adapted
to the case of computing and enumerating {\teaxp}s and {\tecxp}s.

\subsection{Computation of {\teaxp}s \& {\tecxp}s}

\jnoteF{CXps are believed to be \emph{easier} than AXps, but we can
  formulate both as MSMP.}

As noted earlier in the paper, existing algorithms for computing AXps
mimic those for computing MUSes of logic formulas. The same
observation can be made in the case of distance-restricted AXps, i.e.\ any algorithm developed for computing
one MUS of a logic formula can be used for computing one $\eaxp$.
Furthermore, because the predicates $\waxp$, $\wcxp$, $\ewaxp$ and
$\ewcxp$ are monotone, the algorithms used for computing
{\teaxp}s can be used for computing {\tecxp}s.
As a result, throughout this section, we will only detail the
computation of {\teaxp}s.

By double-negating~\eqref{eq:waxpg}, it is plain to conclude that a
set of features $\fml{X}$ is a \tewaxp if the following statement does
\emph{not} hold:
\begin{align}
  \exists&(\mbf{x}\in\mbb{F}).\\
  &\left(\bigwedge\nolimits_{i\in\fml{X}}(x_i=v_i)\land(||\mbf{x}-\mbf{v}||_{p}\le\refd)\right)\land(\kappa(\mbf{x})\not=c)\nonumber
\end{align}
or, alternatively, that the following logic formula is \emph{not}
satisfiable:
\begin{equation}
  \left(\bigwedge\nolimits_{i\in\fml{X}}(x_i=v_i)\land(||\mbf{x}-\mbf{v}||_{p}\le\refd)\right)\land(\kappa(\mbf{x})\not=c)
\end{equation}
which corresponds to deciding for the non-existence of a constrained
\taex when the set of constraints requires the features in $\fml{X}$
to take the values dictated by $\mbf{v}$.

The observations above provide all the insights that justify the
algorithms for distance-restricted explanations proposed in recent
work~(\citealp{barrett-corr22}; \citealp{hms-corr23};
\citealp{barrett-nips23}).
Moreover,
the same observations and the monotonicity of $\ewaxp$,
allow mimicking the algorithms developed for MUS extraction (and also
for the case of computing one AXp) to the case of computing one
\teaxp, provided the oracle used decides the non-existence of an
adversarial example.

For the two algorithms described in this section, calls to the
robustness oracle will be denoted by $\robt$. Furthermore, we will
require that the robustness oracle allows some features to be fixed,
i.e.\ the robustness oracle decides the existence of constrained
adversarial examples. As a result, the call to $\robt$ uses as
arguments the distance $\epsilon$ and the set of fixed features, and
it is parameterized by the explanation problem $\fml{E}$ and the norm $p$
used.


\paragraph{Deletion-based algorithm.}
%
\cref{alg:del} summarizes the simplest algorithm for the computation
of a single \teaxp. (The same algorithm is often used for MUS
extraction, but also for computing distance-unrestricted
AXps/CXps. Since {\teaxp}s and {\tecxp}s are also examples of
MSMP (minimal sets over a monotone
predicate~\cite{msjb-cav13,msjm-aij17}), then the same algorithm can
also be used for computing one \tecxp.)
\begin{algorithm}[t]
  \input{./algs/findaxp_del}
  \caption{Deletion algorithm to find \teaxp
  }
  \label{alg:del}
\end{algorithm}
As can be observed, given some $l_p$ distance $\epsilon>0$, the
algorithm iteratively picks a feature to be allowed to be
unconstrained, starting by fixing all features to the values dictated
by $\mbf{v}$. If no adversarial example is identified, then the
feature is left unconstrained; otherwise, it becomes fixed again.
It is plain that the algorithm requires $\Theta(|\fml{F}|)$ calls to
the robustness oracle. Furthermore, the robustness oracle must enable
the fixing of some or all of the features.
In the context of distance-restricted explanations, variants of the
deletion algorithm were studied in recent
work~(\citealp{barrett-corr22}; \citealp{hms-corr23};
\citealp{barrett-nips23}).

It is convenient to introduce the concept of \emph{transition
feature} $i\in\fml{F}$. Given some set $\fml{S}$ of fixed features,
$i\in\fml{F}$ is a transition feature if (a) when $i$ is not fixed
(i.e.\ $i\not\in\fml{S}$), then an adversarial example exists;
and (b) when $i$ is fixed (i.e.\ $i\in\fml{S}$), then no adversarial
example exists. The point is that $i$ must be included in $\fml{S}$
for $\fml{S}$ to represent a \tewaxp. Accordingly, the deletion
algorithm iteratively decides whether a feature is a transition
feature, given the features already in~$\fml{S}$.


\paragraph{Dichotomic search algorithm.}
%
Aiming to reduce the overall running time of computing one \teaxp,
this paper seeks mechanisms to avoid the $\Theta(|\fml{F}|)$
\emph{sequential} calls to the robustness oracle. For that, it will be
convenient to study another (less used) algorithm, one that implements
dichotomic (or binary) search~\citep{sais-ecai06}.
\begin{algorithm}[t]
  \input{./algs/findaxp_dicho}
  \caption{Dichotomic search algorithm to find \teaxp
  }
  \label{alg:dicho}
\end{algorithm}
The dichotomic search algorithm is shown in~\cref{alg:dicho}.%
\footnote{%
With a slight abuse of notation, set $\fml{W}$ is assumed to be
ordered, such that $\fml{W}_{a..b}$ denotes picking the elements
(i.e.\ features) ordered from index $a$ up to $b$. It is also assumed
that $\fml{W}_{a..b}$, with $a=0\lor{a>b}$ represents an empty set.}
At each iteration of the outer loop, the algorithm uses binary
search in an internal loop to find a \emph{transition feature},
%
i.e.\ fixing the features in $\fml{S}\cup\fml{W}_{1..j}$ does not
yield an adversarial example, but fixing only the features in
$\fml{S}\cup\fml{W}_{1..i-1}$ exhibits an adversarial example, if
$i\not=0$.
(Upon termination of the inner loop, if $i=0$, then
$\fml{W}=\emptyset$.)
Moreover, the features in $\fml{W}_{j+1..{|\fml{W}|}}$ can be safely
discarded.
As a result, it is the case that the inner loop of the algorithm
maintains the following two invariants: (i)
$\ewaxp(\fml{S}\cup\fml{W}_{1..j})$; and (ii)
$(i=0)\lor\neg\ewaxp(\fml{S}\cup\fml{W}_{1..i-1})$. (The pseudo-code
only shows the first invariant.)
Clearly, the updates to $i$ and $j$ in the inner loop maintain the
invariants.
Moreover, it is easy to see that the features in $\fml{S}$ denote a
subset of a \teaxp, since we only add to $\fml{S}$ transition
features; this represents the invariant of the outer loop. When there
are no more features to analyze, then $\fml{S}$ will denote a
\teaxp.
If $k_M$ is the size of the largest \teaxp, then the number of calls
to the robustness oracle is $\fml{O}(k_M\log{m})$.
If the largest \teaxp is significantly smaller than $\fml{F}$, then
one can expect dichotomic search to improve the performance with
respect to the deletion algorithm.

It will be helpful to view the dichotomic search algorithm as a
procedure for analyzing chunks of features. A similar observation can
be made with respect to the QuickXplain~\citep{junker-aaai04} and the
Progression~\citep{msjb-cav13} algorithms. We will later see that
parallelization can be elicited by analyzing different chunks of
features in parallel.

%
%

\paragraph{Other algorithms.}
%
As noted earlier, several other algorithms have been devised over the
years. These include the well-known QuickXplain~\citep{junker-aaai04}
algorithm, the insertion algorithm~\citep{puget-ecai88}, and the
progression algorithm~\citep{msjb-cav13}, among others~\citep{blms-aic12}.
Nevertheless, and despite some past attempts at parallelizing some of
these algorithms~\citep{bmms-sat13}, performance gains have been
modest. Later in the paper, we revisit the dichotomic search
algorithm, and introduce the \swiftxp algorithm in~\cref{sec:nualg}.

In the case of MCSes or CXps, there are more efficient algorithms (in
the number of oracle calls) that can be used~\citep{msm-ijcai20}.
One example is the \emph{clause} $D$ (CLD) algorithm~\citep{mshjpb-ijcai13},
where elements that can be dropped from the minimal set are
iteratively identified, and several can be removed in each oracle
call.
In contrast with algorithms for MUSes and AXps, an MCS (or CXp) can be
decided with a single call using the so-called clause $D$ (or
disjunction clause).
If the elements represented in the clause $D$ represent a minimal set,
then no additional elements can be found, and so the algorithm
terminates by reporting the minimal set~\citep{mshjpb-ijcai13}.
As shown later in the paper, we can use parallelization to emulate the
CLD algorithm in the case of computing one \teaxp, (Evidently, the
same observation also holds true for AXps and MUSes.)%
\footnote{%
It should be underscored that the CLD algorithm has \emph{never} been
used in the past for finding MUSes, AXps, or {\teaxp}s. However, in
the case parallelization is used, the CLD algorithm can be emulated,
as shown later in the paper.}

\jnoteF{Enumeration algorithm.}

\subsection{Additional Problems \& Main Limitations}

\paragraph{Enumeration of explanations.}
The standard solution for the enumeration of explanations is based on
the MARCO algorithm~\citep{lpmms-cj16,msm-ijcai20}. This is the case
with distance-unrestricted explanations~\citep{inams-aiia20}. Moreover,
by changing the oracles used, and in light of duality between
distance-restricted explanations, either the MARCO algorithm, or one
of its variants, can also be used for enumerating distance-restricted
explanations. Observe that enumeration is limited by the number of
explanations and the time for computing each \teaxp or \tecxp.

\paragraph{Main limitations.}
The algorithms outlined in this section, or the examples used in
recent work~\citep{barrett-corr22,hms-corr23,barrett-nips23}, link the
performance of computing explanations to the ability of deciding the
existence of adversarial examples. More efficient tools for deciding
the existence of adversarial examples
(e.g.\ from~\cite{johnson-sttt23}) will result in more efficient
algorithms for computing distance-restricted explanations.
Nevertheless, one bottleneck of the algorithms discussed in this
section is that the number of calls to an oracle deciding the
existence of an adversarial example grows with the number of features.
For complex ML models with a large number of features, the overall
running time can become prohibitive.
The next section outlines novel insights on how to reduce the overall
running time by exploiting opportunities to parallelize calls to the
robustness oracle.

\jnoteF{Main limitation. Proposed algorithms are
  sequential. Parallelization has not been successful before, either
  in MUS/MCS, primes, etc.}

%% file: algs/findaxp_del.tex
\begin{flushleft}
  \hspace*{\algorithmicindent}
  \textbf{Input}: {
    Arguments: 
    $\epsilon$;
    Parameters: 
    $\fml{E}$,
    $p$}\\
  \hspace*{\algorithmicindent}
  \textbf{Output}: {One \teaxp $\fml{S}$}
\end{flushleft}

\begin{algorithmic}[1]
  \Function{$\findaxpdel$}{$\epsilon;\fml{E},p$}
  \State{$\fml{S}\gets\fml{F}$}
  \Comment{Initially, no feature is allowed to change}
  \For{$i\in\fml{F}$}
  \Comment{Invariant: $\ewaxp(\fml{S})$}
  \State{$\fml{S}\gets\fml{S}\setminus\{i\}$}
  \State{$\outc\gets\robt(\epsilon,\fml{S};\fml{E},p)$}
  \If{$\outc$}
  \State{$\fml{S}\gets\fml{S}\cup\{i\}$}
  \EndIf
  \EndFor
  \State{\tbf{return} $\fml{S}$}
  \Comment{$\ewaxp(\fml{S})\,{\land}\,\mathsf{minimal}(\fml{S})\limply\eaxp(\fml{S})$}
  \EndFunction
\end{algorithmic}

%% file: algs/findaxp_dicho.tex
\begin{flushleft}
  \hspace*{\algorithmicindent}
  \textbf{Input}: {
    Arguments: 
    $\epsilon$;
    Parameters: 
    $\fml{E}$,
    $p$}\\
  \hspace*{\algorithmicindent}
  \textbf{Output}: {One \teaxp $\fml{S}$}
\end{flushleft}

\begin{algorithmic}[1]
  \Function{$\findaxpdicho$}{$\epsilon;\fml{E},p$}
  \State{$(\fml{S},\fml{W})\gets(\emptyset,\fml{F})$} 
  \Comment{Precondition: $\ewaxp(\fml{S}\cup\fml{W})$}
  \While{$\fml{W}\not=\emptyset$}
  \Comment{Invariant: $\exists(\fml{X}\in\dsym\mbb{A}).\fml{S}\subseteq\fml{X}$}
  \State{$(i,j)\gets(0,|\fml{W}|)$}
  \While{$i<j$} 
  \Comment{Invariant: $\ewaxp(\fml{S}\cup\fml{W}_{1..j})$}
  \State{$t\gets\lfloor\sfrac{(i+j)}{2}\rfloor$}
  \If{$\robt(\epsilon,\fml{S}\cup\fml{W}_{1..t};\fml{E},p)$} 
  \State{$i\gets{t+1}$}
  \Comment{Fix more features}
  \Else
  \State{$j\gets{t}$}
  \Comment{Free more features}
  \EndIf
  \EndWhile
  \State{$(\fml{S},\fml{W})\gets(\fml{S}\cup\fml{W}_{j..j},\fml{W}_{1..j-1})$}
  \EndWhile
  \State{\tbf{return} $\fml{S}$}
  \Comment{$\exists(\fml{X}\:\!{\in}\:\!\dsym\mbb{A}).(\fml{S}\:\!{\subseteq}\:\!\fml{X})\:\!{\land}\:\!(\fml{W}\:\!{=}\:\!\emptyset)\:\!{\limply}\:\!\eaxp(\fml{S})$}
  \EndFunction
\end{algorithmic}
%
%

%% file: nualg.tex
\section{\swiftxp}
\label{sec:nualg}

\input{nupar}

%

%
\begin{algorithm}[t]
\input{./algs/swiftxp}

  \caption{\swiftxp algorithm to find one \teaxp}
  \label{alg:swiftxp}
\end{algorithm}

\paragraph{A parallel algorithm for finding one \teaxp.}
\cref{alg:swiftxp} outlines the \swiftxp algorithm, which runs in
parallel on multi-core CPU or GPU.
The procedure takes as input the explanation problem $\fml{E}$, an
$l_p$ distance $\epsilon > 0$, a threshold $\delta \in [0, 1]$ used
for activating the optional feature disjunction check, and the number
$q$ of available processors; and returns a \teaxp
$\fml{S}\subseteq\fml{F}$.
Intuitively, the algorithm implements a parallel dichotomic search by
splitting the set of features to analyze into a collection of chunks
and instruments decision oracle calls checking the existence of
adversarial examples, done in parallel on those chunks.
Upon completion of such a parallel oracle call, the algorithm proceeds
by zooming into a chunk that is deemed to contain a transition
feature.

The algorithm starts by initializing the operational set of features
$\fml{W}$ to contain all the features of $\fml{F}$ and a
subset-minimal \teaxp $\fml{S}$ to extract as $\emptyset$.
(Note that one can potentially impose a heuristic feature order on
$\fml{F}$ aiming to quickly remove irrelevant features.)
In each iteration of the outer loop, the lower and upper bounds $\ell$
and $u$ on feature indices are set, respectively, to 1 and $|\fml{W}|$.
%

Each transition element is determined in parallel by the inner loop of
the algorithm, which implements dichotomic search and iterates until
$\ell+1 = u$.
An iteration of this loop splits the set of features $\fml{W}$ into
$\omega$ chunks determined by the splitting indices kept in
$\fml{D}\subseteq\fml{W}$.
(Note that the value of $\omega$ equals either the number of available
CPUs $q$ or the number of remaining features in $\fml{W}$, depending
on which of these values is smaller.)
Given the largest feature index $i\in\fml{D}$ in each such chunk, the
iteration tests whether an adversarial example can be found while
fixing the features $\fml{S}\cup\fml{W}_{1..i}$.
The test is applied in parallel for all the splitting indices
$i\in\fml{D}$ employing $\omega$ CPUs.

\begin{algorithm}[t]

\input{./algs/featD}
  \caption{Parallelized feature disjunction check}
  \label{alg:fd}
\end{algorithm}

%
The aim of the algorithm is to determine the first case when an oracle
call reports that no adversarial example exists, i.e. that
$\oper{AE}_t=\False$ s.t. $\oper{AE}_{\ge t}=\False$ and
$\oper{AE}_{<t}=\True$.
Importantly, as soon as such case $t$ is determined, all the parallel
jobs are terminated.
(Note that in practice terminating the jobs after $i$ s.t.
$\oper{AE}_i=\False$ and before $i$ s.t. $\oper{AE}_i=\True$ helps to
save a significant amount of time spent on the parallel oracle calls.)
The inner loop proceeds by zooming into the $t$'th chunk of features
by updating the values of the lower and upper bounds $\ell$ and $u$ as
it is deemed to contain a transition feature.
If all the oracle calls unanimously decide that an adversarial example
exists (or does not exist), the algorithm proceeds by zooming into the
corresponding \emph{boundary} chunk of features.
Note that if the value of the upper bound $u$ is updated from
$|\fml{W}|$ all the way down to 1, which happens if all the parallel
oracle calls report no adversarial example, the algorithm needs to
check whether set $\fml{S}$ is sufficient for the given prediction.
If it is, the algorithm terminates by reporting $\fml{S}$.
Otherwise, it collects a newly determined transition feature and
proceeds by updating $\fml{W}$ and $\fml{S}$.

Finally, we integrate an analogue of the CLD
procedure~\citep{mshjpb-ijcai13} widely used in the computation of
minimal correction subsets (MCS) of an unsatisfiable CNF formula.
The analogue is referred to as feature disjunction check (see
$\msf{FeatDisjunct}$ in \cref{alg:fd}) and used as an optional
optimization step in \cref{alg:swiftxp} at the beginning of the main
(outer) loop.
We implement a heuristic order over $\fml{F}$ and activate
$\msf{featureDisjunct}$ (given some threshold $\delta$) at the last
iterations of $\findaxpswift$, where it is likely to conclude that all
the features in a selected subset $\fml{T}\subseteq\fml{W}$ of size
$\min(q, |\fml{W}|)$ are relevant for the explanation and can be
safely moved to $\fml{S}$ at once; otherwise, one can randomly pick a
single feature in $\fml{T}$ among those verified as irrelevant
features, i.e.\ removing the feature does not yield an adversarial
example, and make it free.
Finally, we observe that after running $\msf{featureDisjunct}$, the
algorithm does not invoke dichotomic search in the subsequent
iterations.



\input{nrelw}

%% file: nupar.tex
\paragraph{Remarks about parallelization of finding one \teaxp.}
%
As noted earlier in the paper, a limitation of algorithms for
computing AXps, {\teaxp}s, or MUSes is that one element is identified
at each step, and so it is unlikely that one can improve substantially
over algorithms requiring $\fml{O}(m)$ calls to a robustness oracle.

Nevertheless, as one can conclude from inspection of dichotomic
search algorithm (see~\cref{sec:core}), in the worst case, the main
loop runs $\fml{O}(k_M)$ times the second loop and, each time, it
calls the robustness oracle. In the case of the dichotomic search
algorithm, the second loop requires $\fml{O}(\log{m})$ calls to the
robustness oracle. If we were able to reduce the number of robustness
calls of the second loop to one, then the overall number of calls to a
robustness oracle would be $\fml{O}(k_M)$ with a constant of 2.

Suppose that we could run in parallel as many calls to the robustness
oracle as we deemed necessary. Then, we could run $\fml{O}(m)$ calls
to decide which feature should be deemed a transition feature, as
follows.
For processor $r$, we would test the existence of an adversarial
example by fixing the features in
$C_r\triangleq\fml{S}\cup\fml{W}_{1..r}$.
If for the value $r$, $C_r$ does not yield an adversarial
example, and $C_{r-1}$ does yield such an adversarial example, then $r$
represents the transition feature.

Given the algorithm outlined above, and assuming the existence of
arbitrary processors to run the robustness tool in parallel, then we
would be able to find a \teaxp with $\fml{O}(k_M)$ calls to a
robustness oracle. If $k_M\ll{m}$, then this would reduce
(substantially) the number of sequential calls to the robustness
oracle.%
\footnote{%
Unfortunately, AXps can represent a significant percentage (often up
to 65-75\%) of the set of features. As a result, the dichotomic search
algorithm is unlikely to outperform (in terms of robustness oracle
calls) the basic deletion-based algotihm. However, as shown
in~\cref{sec:res}, the sizes of {\teaxp}s can represent a small 
percentage (in some cases down to 15\%) of the number of features; hence
dichotomic search can become more effective.}
Furthermore, in practice, the number of features can be much larger
than the number of available processors, and so we need to devise
mechanisms to parallelize the internal loop of the dichotomic search
algorithm, without using too many processors.

We can view the dichotomic search algorithm as a procedure for
analyzing (possibly in parallel) different \emph{chunks} of features,
with the purpose of finding an element that \emph{must} be included in
some minimal set.
Perhaps unsurprisingly, both the QuickXplain and the Progression
algorithm can also be viewed as procedures for analyzing different
chunks of features.

Moreover, and as hinted in~\cref{sec:core}, in the case of {\teaxp}s
(and also AXps and MUSes) we can emulate the CLD algorithm with many
parallel oracle calls.
As before, we start by assuming that we have enough available
processors. Suppose that $\fml{F}$ is partitioned into three sets.
$\fml{S}$ denotes the elements that are known to be included in some
minimal set, $\fml{N}$ denotes the elements that have been discarded
from being included in some minimal set (i.e.\ these elements will no
longer be used), and $\fml{W}$ denotes the elements we are unsure
about. Let $q=|\fml{W}|$.
Create $q$ sets of fixed features: $\fml{S}\cup\fml{W}\setminus\{i\}$,
for each $i\in\fml{W}$. In addition, run in parallel each of possible
sets of fixed features, using $q$ processors.
If $\fml{S}\cup\fml{W}$ represents a minimal set,
e.g.\ a \teaxp or an AXp or an MUS, then all of the $q$ calls will
return an indication that some adversarial example was found. Thus,
with a single step (involving $q$ parallel oracle calls) we are able
to decide that the target set $\fml{S}\cup\fml{W}$ represents a
minimal set. In this case, all the features in $\fml{W}$ are added to
$\fml{S}$ and the algorithm terminates by returning $\fml{S}$.
Furthermore, if for some feature $i$, the oracle call decides that no
adversarial example exists, then feature $i$ can be dropped from
$\fml{W}$ and added to $\fml{N}$. If no adversarial example is
returned for multiple features, only one of the features can be
dropped.
In practice, implementing the CLD algorithm in parallel may require
multiple sequential calls, concretely when the number of available
processors is insufficient.

It should be noted that the parallelization of dichotomic search and
of CLD serve different purposes, and so one must be able to pick the
right parallel oracle calls to make.
At each step, one can either mimic the dichotomic search algorithm to
drop multiple features, or mimic the CLD algorithm.
For example, one may start by running several steps of the (parallel)
dichotomic search algorithm, and at some point switch to the
(parallel) clause $D$ algorithm.
The rest of this section addresses how this can be done in practice.

%% file: algs/swiftxp.tex
\begin{flushleft}
  \hspace*{\algorithmicindent}
  \textbf{Input}: {
    Arguments:   $\epsilon$, $q$, $\delta$;
    Parameters: $\fml{E}$,  $p$}\\
\hspace*{\algorithmicindent}
\textbf{Output}: {One \teaxp $\fml{S}$}
\end{flushleft}
\begin{algorithmic}[1]
  \Function{$\findaxpswift$}{$\epsilon, q, \delta;\fml{E},p$}
  \State{$(\fml{W}, \fml{S}) \gets (\fml{F},  \emptyset)$}
  \Comment{Precond: $\ewaxp(\fml{F})\land(q\ge2)$}
  \While{$\fml{W} \neq \emptyset$}
       \Comment{$\fml{S}\subseteq\fml{X}\in\dsym\mbb{A}$}
       \If{ $|\fml{W}| < \delta\times |\fml{F}|$ }
       	  \Comment{Run FD check}
          \State{$(\fml{W},\fml{S})\gets\msf{FeatDisjunct}(\epsilon, q, \fml{W}, \fml{S}; \fml{E},p)$}
          \State{\bfseries{continue}}
       \EndIf
      \State{$(\ell,u)\gets(0,|\fml{W}|)$}
      \While{$\ell+1<u$}
           \Comment{Inv.: $\ewaxp(\fml{S}\cup\fml{W}_{1..u})$}
           \State{$\omega\gets\min(q, u-\ell)$}
           \Comment{\# parallel calls}
           \State{$\sigma\gets\lfloor\sfrac{(u-\ell)}{\omega}\rfloor$}
           \Comment{$\sigma$: chunk size}
           \State{$\fml{D}\gets\left\{\ell+\iota\times\sigma\;|\;\iota\in\{1,\ldots,\omega\}\right\}$}
           \algrenewcommand\algorithmicdo{\textbf{do in parallel}}
           \For{$i \in \fml{D}$}
           \State{$\oper{AE}_i\;\!{\gets}\;\!\robt(\epsilon,\fml{S}\;\!{\cup}\;\!\fml{W}_{1..i};\fml{E},p)$}
           \EndFor
           \algrenewcommand\algorithmicdo{\textbf{do}}
           \State{$u\gets\min(\{i\in\fml{D}\mid\oper{AE}_i=\False\}\cup\{u\})$}
           \State{$\ell\gets\max(\{i\in\fml{D}\mid i<u\}\cup\{\ell\})$}
       \EndWhile
       \If{${u=1}\land\neg\robt(\epsilon,\fml{S};\fml{E},p)$} 
       \State{\Return{$\fml{S}$}}
       \EndIf
       \State{$(\fml{W}, \fml{S}) \gets (\fml{W}_{1..u-1}, \fml{S}\cup\fml{W}_{u..u})$}
       \EndWhile
  \State\Return{$\fml{S}$}
\EndFunction
\end{algorithmic}

%% file: algs/featD.tex
\begin{flushleft}
  \hspace*{\algorithmicindent}
  \textbf{Input}: {
    Arguments:   $\epsilon$, $\fml{W}$, $\fml{S}$;
    Parameters: $\fml{E}$,  $p$}\\
\end{flushleft}
\begin{algorithmic}[1]
  \Procedure{$\msf{FeatDisjunct}$}{$\epsilon, q, \fml{W}, \fml{S}; \fml{E}, p$}
  \State{$\fml{T}\gets\fml{W}_{\max(|\fml{W}|-q+1,1)..|\fml{W}|}$}
           \algrenewcommand\algorithmicdo{\textbf{do in parallel}}
           \For{$i \in \fml{T} $}
               \Comment{Traversing target set $\fml{T}$}
               \State{$\oper{AE}_i\gets\robt(\epsilon,\fml{S}\cup\fml{W}\setminus\{i\};\fml{E},p)$}
           \EndFor
           \algrenewcommand\algorithmicdo{\textbf{do}}
       \If{$\bigland\nolimits_{i \in \fml{T}} \oper{AE}_i = \True$}
           \Comment{Fix more features}
           \State\Return{$(\fml{W}\setminus \fml{T}, \fml{S}\cup \fml{T})$}
       \Else
           \Comment{Free one feature}
           \State{$j\gets\msf{PickRandom}(\{ i\in\fml{T} \mid \oper{AE}_i = \False \})$}
           \State\Return{($\fml{W}\setminus \{j\},\fml{S})$}
       \EndIf
\EndProcedure
\end{algorithmic}

%% file: nrelw.tex
\paragraph{Related work.}
Previous work on {\teaxp}s considered sequential
algorithms~(\citealp{barrett-corr22}; \citealp{hms-corr23};
\citealp{barrett-nips23}). 
The same holds true for earlier work on computing AXps.
In the more general setting of model-based diagnosis (MBD), existing
work on parallelization considers solely local optimizations to 
sequential algorithms~(\citealp{abreu-epia13};
\citealp{jannach-aaai15}; \citealp{jannach-jair16};
\citealp{jannach-hpcr18}; \citealp{felfernig-ismis20};
\citealp{felfernig-aaai23}).
For finding one MUS,~\citep{bmms-sat13} proposes the parallelization
of the deletion-based algorithm.
This work is the closest to the ideas developed in this paper, with
significant differences, including the parallelization of the
dichotomic search algorithm, and the parallelization of the CLD
algorithm.
%

\jnoteF{Move the previous paragraph to the end of this section!}

%% file: res.tex
\input{./figs/swiftxp_vs_del}

\section{Experimental Evidence} \label{sec:res}
We assess our approach to computing \teaxp for DNNs on well known
image data 
and compare with the SOTA  deletion algorithm.
Additional results on comparison with dichotomic search can be
found in the supplementary material.

\paragraph{Experimental Setup.}
All experiments were carried out on a high-performance computer
cluster with machines equipped with AMD EPYC 7713 processors.
Each instance test is provided with 2 and 60 cores, resp., when running
the sequential algorithms deletion (\cref{alg:del}), dichotomic (\cref{alg:dicho}),
and our parallel \swiftxp algorithm, namely 1 core for 1 oracle used
and 1 additional core to run the main script.
Furthermore, the memory limit was set to 16GB, and the time limit
to 14400 seconds (i.e.\ 4 hours).

\paragraph{Prototype Implementation.}
The proposed approach was prototyped as a set of Python scripts\footnote{%
The sources will be made publicly available with the paper's final version.
}, %
and PyTorch library \citep{pytorch-nips19} was used to  train and
handle the learned DNNs.
A unified Python interface for  robustness oracles is implemented and it
enables us to use any DNN reasoner of the VNN-COM~\citep{johnson-sttt23}.
MN-BaB~\citep{Vechev-iclr22}, which is a complete neural network
verifier, is used to instrument AEx checking on CPU mode. (Note that
one can run MN-BaB on (multi-)GPU with CUDA library to boost/speed up
the oracle calls, but this clearly would not change the results of our
evaluation.)
Moreover, Gurobi~\citep{gurobi} is applied for empowering MN-BaB resolution.
Other tools like Marabou~\citep{katz2019marabou} were also tested, but
were unable to scale for large benchmarks, thus we focus solely
on the results obtained with MN-BaB.
Besides, we implemented  the {\it pixel sensitivity} ranking heuristic,
as described in~\citep{barrett-nips23}, and LIME \citep{guestrin-kdd16}
for the traversal order of features in all algorithms outlined above.
Note that preliminary results shown that traversal heuristic based
on sensitivity score is faster to compute and often provide  smaller
explanations\footnote{This remark joins the observation of \citep{barrett-nips23}
of the empirical assessment of explanations size without and with
sensitivity heuristic traversal order.}.
Therefore, presented results in this section are showing
experiments applying pixel sensitivity heuristic.

\paragraph{Data and ML  Models.}
The experiments focus on two well-known image datasets, that
have been studied in~\citep{barrett-nips23}.
Namely, we evaluate the widely used {\it MNIST}~\citep{Li-spm12,PaszkeGMLBCKLGA19}
dataset, which features hand-written digits from 0 to 9.
Also, we consider the image dataset {\it GSTRB}~\citep{StallkampSSI12} of traffic signs, and
we select a collection of training data that represents the top 10 classes of
the entire data.
We considered different $\epsilon$ values for each model and image size, following
the experimental protocol of \citep{barrett-nips23}, e.g. for {MNIST} benchmark
$\epsilon$  varies from $0.003$ to $0.15$.
Besides, the parameter $\delta$ in \swiftxp is set to 0.75
for all benchmarks.
%

\paragraph{Results.}
\cref{tab:swift-vs-del} summarizes the results comparing \swiftxp and
the baseline linear (deletion-based) method on fully-connected (dense)
and convolutional NNs  trained with the above image datasets.
As can be observed from \cref{tab:swift-vs-del}, \swiftxp
significantly surpasses the deletion algorithm on all tested
benchmarks.
More specifically, \swiftxp is on average up to 4 times faster than
deletion (e.g.\ 4.5 and 4 times faster on {\it gtsrb-dense} and {\it
gtsrb-convSmall}), with a few exceptions on smaller image inputs but
still takes the advantage on all computed \teaxp's (i.e.\ $2.6$ and
$1.1$ faster on {\it mnist-denseSmall} and {\it mnist-dense}).
Moreover, the linear approach fails to produce a \teaxp on larger
(convolutional) models for all tested image samples within 4-hours
time limit, whilst \swiftxp successfully finds an explanation on all
tests with an average runtime of  $4449.4$ sec ($\sim$74 min) and
$5132.8$ sec ($\sim$85 min), resp., on convolutional {\it gtsrb} and
{\it mnist} NNs.
Interestingly, the largest runtimes reported for \swiftxp are smaller
than or close to the smallest runtimes of the deletion
algorithm.
This observation does not come as a surprise if we recall that the
number of robustness oracle calls is drastically reduced in \swiftxp
when compared to the linear approach, which is always exactly the
number of data features.
(Note that in case of dichotomic  search, as shown in our experiments,
this number is always larger than the image size.)
For instance in {\it GTSRB}, \swiftxp makes on average between 13.8\%
to  22.7\% ($q$ parallel) oracle calls of the total number of calls
instrumented by the deletion algorithm (i.e.\ 77.3\% to 86.2\%  less).
Focusing solely on the performance of \swiftxp, we observe that
activation of the feature disjunction (FD) technique is more effective
when the feature set size $|\fml{W}|$ left to inspect is smaller than
the average size of \teaxp.
Also, we observe that deactivating FD in the first iterations of
\swiftxp enables us to drop  up to 50\% of (chunks of) features with a
few iterations in the inner loop. 
Furthermore, one can see from \cref{tab:swift-vs-del} that the average
success of FD to capture $q$ transition features in one iteration
varies from $33.2\%$ to $77.5\%$ for {\it GTSRB} and $11.5\%$ to
$21.3\%$ for {\it MNIST}.
Also, the larger \teaxp is, the more likely it is that FD detects
chunks of $|\fml{T}|$ transition features to augment $\fml{S}$ with.
Finally, we note that when comparing sensitivity and LIME feature
traversal heuristics, the former one improves the effectiveness of FD,
and thus helps us reduce the total number of (parallel) calls.

\paragraph{Discussion of performances and potential improvements.}
To sum up, these observations let us conclude that \cref{alg:fd} and
\cref{alg:swiftxp} complement each other, such that when the input
data or a target \teaxp is expected to be large (which can be
heuristically measured based on feature/pixel importance score) then 
FD serves to augment $\fml{S}$ by means
of fewer iterations; conversely for smaller $\epsilon$ distance or
when a smaller \teaxp is expected, \cref{alg:swiftxp} allows us to
eliminate more features with fewer iterations.
Independently from the outperformance of our approach 
against the SOTA methods demonstrated in this evaluation, 
we observed some (technical) limitations that we are willing to investigate 
in the future. 
First, the prototype does not yet enable multi-core (CPU/GPU) use 
in the oracle; second, it does not support the incremental mode when   
calling iteratively the oracle, which would likely improve the runtimes 
since incremental resolution has been shown 
beneficial in other settings like SAT oracles; third, we are interested 
in analyzing the evolution of the runtime curve when increasing the number 
of processors to hundreds or thousand CPUs/GPUs on data with not much 
larger in number of features.

%% file: figs/swiftxp_vs_del.tex
\sisetup{parse-numbers=false,detect-all,mode=text}
\setlength{\tabcolsep}{4pt}

\begin{table*}[ht]
\centering
\resizebox{\textwidth}{!}{
  \begin{tabular}{l
  S[table-format=1.2]S[table-format=4]
  S[table-format=3]S[table-format=3.1]S[table-format=3.1]S[table-format=3.1]S[table-format=3]
  S[table-format=2.2]S[table-format=3]S[table-format=3]
  S[table-format=2.1]S[table-format=4.1]S[table-format=4.1]S[table-format=4.1]}
\toprule[1.2pt]
\multirow{2}{*}{\bf Model}  & 
\multicolumn{7}{c}{\bf Deletion} & \multicolumn{7}{c}{\bf  \swiftxp}  \\
  \cmidrule[0.8pt](lr{.75em}){2-8}
  \cmidrule[0.8pt](lr{.75em}){9-15}
&  {\bf avgC}  & {\bf nCalls} & {\bf Len} & {\bf Mn}  & {\bf Mx} & {\bf avg } &  {\bf TO } &
		{\bf avgC}  &  {\bf nCalls} & {\bf Len} & {\bf FD\%} & {\bf Mn}  & {\bf Mx} & {\bf avg}  \\
\toprule[1.2pt]

gtsrb-dense  & 0.06 & 1024 & 448 & 52.0 & 76.3 & 63.1 & 0 & 0.23 & 54 & 447 & 77.4 & 10.8 & 14.0 & 12.2 \\
gtsrb-convSmall  & 0.06 & 1024 & 309 & 59.2 & 82.6 & 65.1 & 0 & 0.22 & 74 & 313 & 39.7 & 15.1 & 19.5 & 16.2 \\
gtsrb-conv  & $\textemdash$ & $\textemdash$ & $\textemdash$ & $\textemdash$ & $\textemdash$ & $\textemdash$ & 100 & 96.49 & 45 & 174 & 33.2 & 3858.7 & 6427.7 & 4449.4 \\
mnist-denseSmall  & 0.28 & 784 & 177 & 190.9 & 420.3 & 220.4 & 0 & 0.77 & 111 & 180 & 15.5 & 77.6 & 104.4 & 85.1 \\
mnist-dense  & 0.19 & 784 & 231 & 138.1 & 179.9 & 150.6 & 0 & 0.75 & 183 & 229 & 11.5 & 130.1 & 145.5 & 136.8 \\
mnist-convSmall  & $\textemdash$ & $\textemdash$ & $\textemdash$ & $\textemdash$ & $\textemdash$ & $\textemdash$ & 100 & 98.56 & 52 & 116 & 21.3 & 4115.2 & 6858.3 & 5132.8 \\

\bottomrule[1.2pt]
\end{tabular}
}
\caption{%
\footnotesize{
  Detailed performance evaluation of computing \teaxp for DNNs with 
  \swiftxp and comparison with the baseline linear search (deletion) algorithm. 
%
%
The table shows results for 2 image data, i.e.\  {\it MNIST} (image size: $28\times 28$)  
and {\it GSTRB} (image size: $32\times 32$), and for each row (DNN) a batch of 25 
input images, randomly selected, were tested. 
Columns  {\bf avgC} and  {\bf nCalls}  report, resp.\, the average time 
and average number of instrumented (AEx robustness) oracle  calls.
Column {\bf avg} (resp.\ {\bf Mn} and {\bf Mx} )  reports the average  
(resp.\ min and max) time in seconds to deliver a \teaxp, and 
{\bf TO} is the percentage of timeout tests. 
Lastly, column {\bf Len} reports the average explanation 
length and {\bf FD\%} is the average percentage of successful FD 
calls to augment $\fml{S}$ with $q$ features in one iteration. 
 } }
\label{tab:swift-vs-del}
\vspace{-0.4cm}
\end{table*}

%% file: conc.tex
\section{Conclusions}
\label{sec:conc}

Explainability is a mainstay of trustworthy AI.
Most methods of explainability lack rigor, and this represents a
critical limitation in high-risk and safety-critical uses of ML.
Formal explainability offers the strongest guarantees of rigor, but
its limitations include the complexity of reasoning. This is
demonstrated for example when explaining medium- to large-scale NNs. 
Recent work proposed distance-restricted explanations, which can be
computed by resorting to practically efficient tools for deciding
adversarial
robustness~(\citealp{barrett-corr22}; \citealp{hms-corr23};
\citealp{barrett-nips23}).
The limitations of past work on computing distance-restricted
explanations include understanding and proving some of its key
properties, but also addressing the performance bottleneck represented
by large numbers of features in complex ML models.

This paper provides a detailed formalization of distance-restricted
explanations, showing the existence of key connections with past work
on formal explainability. Furthermore, the paper proposes novel
algorithms for computing distance-restricted explanations for ML
models containing a large number of features. Preliminary experimental
results confirm that significant performance gains can be achieved by
parallelizing the calls to the adversarial robustness oracle.
The obtained results also show indirect promise for improving the
performance of computing MUSes of large logic formulas.
Future work will seek better heuristics for scaling
distance-restricted explainability for highly complex ML models, by
also enabling the answering of a number of relevant explainability
queries~(\citealp{barcelo-nips20}, \citealp{hiims-kr21}; \citealp{marquis-kr21};
\citealp{hims-aaai23}; \citealp{hcmpms-tacas23}). 

%% file: ack.tex
\section{Acknowledgments}
This work was supported by the National Research Foundation, 
Prime Minister’s Office, Singapore under its Campus for Research 
Excellence and Technological Enterprise (CREATE) programme.
The computational work for this article was performed on resources 
of the National Supercomputing Centre, Singapore (\url{https://www.nscc.sg}).

%% file: replbib.tex
\newtoggle{mkbbl}

%% file: togbbl.tex
\settoggle{mkbbl}{false}

%% file: appdx.tex

\input{proofs}

\input{example-plot}

\section{Additional Results}

Detailed results of  \swiftxp and dichotomic search 
assessment are reported in, respectively, in \cref{tab:swiftxp} and 
\cref{tab:dicho}.

\input{./figs/swiftxp}

\input{./figs/dicho}

%


%% file: proofs.tex
\section{Proofs of Propositions}

\begin{proposition} [Proposition 5 in the paper]
\label{prop:dwaxp_epsilon_eta}
  Let $\fml{E}$ represent some explanation problem.
  Let $0<\epsilon<\eta$ and $p$ some norm. For any
  $\fml{S}\subseteq\fml{F}$:
  \begin{enumerate}[nosep]
  \item If $\ewaxp(\fml{S};\fml{E},\eta,p)$, then
    $\ewaxp(\fml{S};\fml{E},\epsilon,p)$;
  \item If $\ewcxp(\fml{S};\fml{E},\epsilon,p)$, then
    $\ewcxp(\fml{S};\fml{E},\eta,p)$;
  \end{enumerate}
\end{proposition}
\begin{proof}
   
  Consider $\fml{E}$ some explanation problem, $0<\epsilon<\eta$, $p$ some norm:
   
  \begin{description}
  
  \item [Statement 1.]
  Let  $\fml{S}\subseteq\fml{F}$ be a subset of features such that $\ewaxp(\fml{S};\fml{E},\eta,p)$ is verified.
  
  Suppose by contradiction that $\ewaxp(\fml{S};\fml{E},\epsilon,p)$ is not verified, then there must exist at least one point $\mbf{v}_{AE}$ such that $\mbf{v}_{AE}$ is an adversarial example, and verifies $\bigwedge\nolimits_{i\in\fml{S}}({v_{AE}}_i=v_i)$.
  
  Since $\mbf{v}_{AE}$ is an adversarial example, then $\lVert\mbf{v}_{AE}-\mbf{v}\rVert_p\le\epsilon$ and $\kappa(\mbf{v}_{AE})\not=c$.
  Because $0<\epsilon<\eta$, then we also have that $\lVert\mbf{v}_{AE}-\mbf{v}\rVert_p\le\epsilon<\eta$, which contradicts the fact that $\ewaxp(\fml{S};\fml{E},\eta,p)$ is verified.
  Thus $\ewaxp(\fml{S};\fml{E},\epsilon,p)$ is also verified.

  \item [Statement 2.]
  Let $\fml{S}\subseteq\fml{F}$ be a subset of features such that $\ewcxp(\fml{S};\fml{E},\epsilon,p)$ is verified, then there exists at least one point $\mbf{v}_{AE}$ such that $\mbf{v}_{AE}$ is an adversarial example, and verifies that $\bigwedge\nolimits_{i\in\fml{F}\setminus\fml{S}}({v_{AE}}_i=v_i)$.
  Since $\mbf{v}_{AE}$ is an adversarial example, then $\lVert\mbf{v}_{AE}-\mbf{v}\rVert_p\le\epsilon$ and $\kappa(\mbf{v}_{AE})\not=c$, and because $0<\epsilon<\eta$, we also have that $\lVert\mbf{v}_{AE}-\mbf{v}\rVert_p\le\eta$.
  Thus $\mbf{v}_{AE}$ verifies that $\bigwedge\nolimits_{i\in\fml{F}\setminus\fml{S}}({v_{AE}}_i=v_i)$, $\lVert\mbf{v}_{AE}-\mbf{v}\rVert_p\le\eta$, and $\kappa(\mbf{v}_{AE})\not=c$, which means that $\ewcxp(\fml{S};\fml{E},\eta,p)$ is verified.

  \end{description}
\end{proof}

%% file: example-plot.tex
\section{Example plots}

\begin{example}
    (Example 7 in the paper)

    Consider a classifier $\fml{M}$ defined on $\fml{F} = \{1, 2, 3\}$ and $\fml{K}=\{0, 1\}$,
    with feature domains $\mbb{D}_1=\mbb{D}_3=\{-0.5,0,0.5,1\}$, and $\mbb{D}_2=\{0,0.5,1\}$.
    Let $\kappa(x_1,x_2,x_3)$ be its classification function
    such that $\kappa(0.5,0.5,1) = 0$, $\kappa(1,0.5,0.5) = 0$, $\kappa(-0.5,1,1) = 0$ and $\kappa(1,1,-0.5) = 0$.
    For any other point $\mbf{x}$, we have $\kappa(\mbf{x})=1$.
    Let $\fml{E} = (\fml{M}, ((1,1,1),1))$.
    Let us use $l_1$, and suppose $\epsilon_1 = 1$ and $\epsilon_2 = 1.5$.
    For $\epsilon_1=1$, there are two AEx $\{(0.5,0.5,1), (1,0.5,0.5)\}$, from which we deduce
    $\dsym\mbb{C}(\fml{E},\refd_1;1) = \{\{1,2\}, \{2,3\}\}$.
    For $\epsilon_2=1.5$, there are four AEx $\{(0.5,0.5,1), (1,0.5,0.5), (-0.5,1,1), (1,1,-0.5)\}$, but then we can deduce
    $\dsym\mbb{C}(\fml{E},\refd_2;1) = \{\{1\}, \{3\}\}$.
    By MHS, we have $\dsym\mbb{A}(\fml{E},\refd_1;1) = \{\{2\}, \{1,3\}\}$,
    but $\dsym\mbb{A}(\fml{E},\refd_2;1) = \{\{1,3\}\}$.
    Clearly, $\{2\} \not \in \dsym\mbb{A}(\fml{E},\refd_2;1)$.
    \end{example}

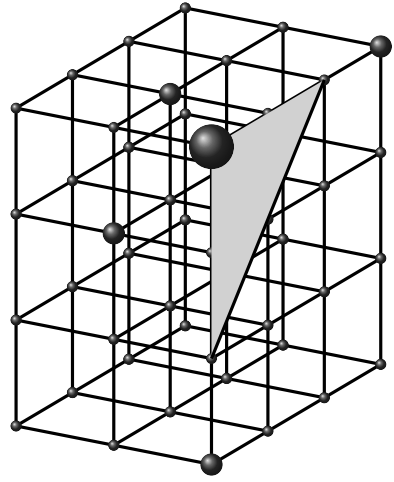
\begin{figure}[h]
    \centering

\tdplotsetmaincoords{70}{120} 
\tdplotsetrotatedcoords{0}{0}{0} 

\begin{tikzpicture}[scale=3,tdplot_rotated_coords,
  rotated axis/.style={->,purple,ultra thick},
  blackBall/.style={ball color = black!80},
  borderBall/.style={ball color = white,opacity=.25}, 
  very thick]

\foreach \x in {-0.5, 0, 0.5, 1}
\foreach \y in {0, 0.5, 1}
\foreach \z in {-0.5, 0, 0.5, 1}{
\ifthenelse{  \lengthtest{\x pt < 1pt}  }{
\draw (\x,\y,\z) -- (\x+0.5,\y,\z);
\shade[rotated axis,blackBall] (\x,\y,\z) circle (0.025cm); 
}{}
\ifthenelse{  \lengthtest{\y pt < 1pt}  }{
\draw (\x,\y,\z) -- (\x,\y+0.5,\z);
\shade[rotated axis,blackBall] (\x,\y,\z) circle (0.025cm);
}{}
\ifthenelse{  \lengthtest{\z pt < 1pt}  }{
\draw (\x,\y,\z) -- (\x,\y,\z+0.5);
\shade[rotated axis,blackBall] (\x,\y,\z) circle (0.025cm);
}{}

}

\foreach \p in {(0.5,0.5,1), (1,0.5,0.5), (-0.5,1,1), (1,1,-0.5)}
\shade[rotated axis,blackBall] \p circle (0.05cm);

\fill[gray!30, opacity=0.9] (1,1,1) -- (1,1,0) -- (0,1,1) -- cycle;
\draw[black, opacity=1] (1,1,0) -- (0,1,1);

\shade[rotated axis,blackBall,color=red] (1,1,1) circle (0.10cm); 

\end{tikzpicture}

\caption{Example 7 plot.}
\label{fig:example-plot}
\end{figure}

In order to illustrate the previous example, we use the plot in Figure~\ref{fig:example-plot}.
The dots in the figure represent all the domain points of the classifier.
The largest dot represents point (1,1,1), with classification value of $1$.
The medium sized dots represent the points $\mbf{x}$ of the domain 
which are classified as $\kappa(\mbf{x})=0$.
The smallest sized dots represent the remaining points of the domain, 
which are classified as $1$.

The shadded triangle represents the sub-space at a $l_1$ norm distance $1$ from the point (1,1,1) 
while fixing the value of feature~2.
Any point $\mbf{x}$ of the domain inside the triangle is classified as $\kappa(\mbf{x})=1$,
thus $\{2\}$ belongs to $\dsym\mbb{A}(\fml{E},\refd_1;1)$.

Observe that if we increase the sub-space to a ($l_1$ norm) distance $1.5$ from the point (1,1,1),
then the new shadded area would include the points in the bottom and right corners,
which are classified as $0$, thus $\{2\}$ will no longer be a $\eaxp$.



%% file: figs/swiftxp.tex
\sisetup{parse-numbers=false,detect-all,mode=text}
\setlength{\tabcolsep}{3pt}

\begin{table}[ht]
\centering
\resizebox{\columnwidth}{!}{
  \begin{tabular}{l
  S[table-format=2.2]S[table-format=3]
  S[table-format=3]S[table-format=2]
  S[table-format=3]S[table-format=4.1]S[table-format=4.1]S[table-format=4.1]}
\toprule[1.2pt]
\multirow{2}{*}{\bf Model}  &  \multicolumn{2}{c}{\bf  AEx} &  \multicolumn{2}{c}{\bf  FD} &  \multicolumn{4}{c}{\bf  \teaxp}  \\
  \cmidrule[0.8pt](lr{.75em}){2-3}
  \cmidrule[0.8pt](lr{.75em}){4-5}
  \cmidrule[0.8pt](lr{.75em}){6-9}
&  {\bf avgC}  &  {\bf nCalls} & {\bf nRuns} &  {\bf S} & {\bf Len}  & {\bf Mn}  & {\bf Mx} & {\bf avg}  \\
\toprule[1.2pt]

gtsrb-dense  & 0.23 & 54 & 53 & 41 & 447 & 10.8 & 14.0 & 12.2 \\
gtsrb-convSmall  & 0.22 & 74 & 73 & 29 & 313 & 15.1 & 19.5 & 16.2 \\
gtsrb-conv  & 96.49 & 45 & 44 & 14 & 174 & 3858.7 & 6427.7 & 4449.4 \\
mnist-denseSmall  & 0.77 & 111 & 110 & 17 & 180 & 77.6 & 104.4 & 85.1 \\
mnist-dense  & 0.75 & 183 & 182 & 21 & 229 & 130.1 & 145.5 & 136.8 \\
mnist-convSmall  & 98.56 & 52 & 51 & 10 & 116 & 4115.2 & 6858.3 & 5132.8 \\

\bottomrule[1.2pt]
\end{tabular}
}
\caption{%
  Detailed performance evaluation of computing \teaxp for DNNs with 
  \swiftxp. 
%
%
%
%
Columns  {\bf avgC} and  {\bf nCalls}  report, resp.\, the average time 
and average number of instrumented  oracle (AEx robustness)  calls.
Columns {\bf nRuns}  and  {\bf S} report the avg.\ total number 
of times we ran FD procedure and the number of successful parallel calls 
that returned all SAT, for computing one \teaxp, s.t. the activation 
parameter $\delta$ is fixed to $0.75$. 
Column {\bf avg} (resp.\ {\bf Mn} and {\bf Mx} )  reports the average time 
(resp.\ min and max) to deliver a \teaxp and {\bf Len} reports the average 
length of the \teaxp's.
}
\label{tab:swiftxp}
\vspace{-0.8cm}
\end{table}

%% file: figs/dicho.tex
\sisetup{parse-numbers=false,detect-all,mode=text}
\setlength{\tabcolsep}{3pt}

\begin{table}[htb]
\centering
\resizebox{\columnwidth}{!}{
  \begin{tabular}{l S[table-format=1.4]
  S[table-format=1.2]S[table-format=4]
  S[table-format=3]S[table-format=3.1]S[table-format=3.1]S[table-format=3.1]S[table-format=3] }
\toprule[1.2pt]
\multirow{2}{*}{\bf Model}  & \multirow{2}{*}{$\epsilon_\infty$} & \multicolumn{2}{c}{\bf AEx } & \multicolumn{5}{c}{\bf \teaxp}  \\
  \cmidrule[0.8pt](lr{.75em}){3-4}
  \cmidrule[0.8pt](lr{.75em}){5-9}
& &  {\bf avgC}  & {\bf nCalls} & {\bf Len} & {\bf Mn}  & {\bf Mx} & {\bf avg } &  {\bf TO }  \\
\toprule[1.2pt]

gtsrb-dense  & 0.02 & 0.09 & 3088 & 448 & 229.0 & 348.8 & 276.2 & 0  \\
gtsrb-convSmall  & 0.003 & 0.12 & 1976 & 309 & 181.9 & 398.2 & 242.8 & 0  \\
gtsrb-conv  &  0.003 &  $\textemdash$ & $\textemdash$ & $\textemdash$ & $\textemdash$ & $\textemdash$ & $\textemdash$ & 100  \\
mnist-denseSmall  &  0.025 &  0.10 & 1104 & 177 & 102.4 & 151.0 & 117.3 & 0  \\
mnist-dense  &  0.08 &  0.13 & 1540 & 231 & 166.7 & 241.9 & 198.4 & 0  \\
mnist-convSmall  &  0.08 &  $\textemdash$ & $\textemdash$ & $\textemdash$ & $\textemdash$ & $\textemdash$ & $\textemdash$ & 100  \\

\bottomrule[1.2pt]
\end{tabular}
}
\caption{%
  Detailed performance evaluation of computing \teaxp for DNNs with 
   dichotomic search algorithm. 
%
%
%
(All columns keep the same meaning as in \cref{tab:swiftxp})
}
\label{tab:dicho}
\vspace{-0.4cm}
\end{table}